# A Fuzzy MLP Approach for Non-linear Pattern Classification

Tirtharaj Dash, H.S. Behera
Department of Computer Science and Engineering
Veer Surendra Sai University of Technology (VSSUT), Burla, Odisha 768018, India
Email: tirtharajnist446@gmail.com, hsbehera_india@yahoo.com

*Abstract:* In case of decision making problems, classification of pattern is a complex and crucial task. Pattern classification using multilayer perceptron (MLP) trained with back propagation learning becomes much complex with increase in number of layers, number of nodes and number of epochs and ultimate increases computational time [31]. In this paper, an attempt has been made to use fuzzy MLP and its learning algorithm for pattern classification. The time and space complexities of the algorithm have been analyzed. A training performance comparison has been carried out between MLP and the proposed fuzzy-MLP model by considering six cases. Results are noted against different learning rates ($\alpha$) ranging from 0 to 1. A new performance evaluation factor 'convergence gain' has been introduced. It is observed that the number of epochs drastically reduced and performance increased compared to MLP. The average and minimum gain has been found to be 93% and 75% respectively. The best gain is found to be 95% and is obtained by setting the learning rate to 0.55.

## 1. Introduction

**P**attern **C**lassification has been one of the most frequently needed tasks in different fields of science and engineering and in many other fields [1]. In real world, every object has certain set of attributes. These attributes help in keeping that object to a certain group, called *class*. A particular group follows a certain pattern. A pattern classification problem can be defined as a problem where an object is assigned into a predefined group or class based on a set of observed attributes related to that object [1]. Let us take a set of examples to support this statement. We classify the programming languages as high-level or low-level, object-oriented or object-based depending on a set of properties possessed by these languages. Several problems in business, science, industry, and medicine can be treated as classification problems. Examples include bankruptcy prediction [2,3], stock market forecasting [2,3], credit scoring, medical diagnosis [4,5], medical data classification [4,5], power quality control and classification [6], handwritten character recognition [7], signature verification [8,9], fingerprint recognition [10,11], speech recognition [12] etc.

Finding solutions to such classification problems has been a crucial research area in the field of technology. Recent works have been done using statistical techniques or using data mining techniques, such as neural computing. Statistical methods use probability theory, decision theory [13] to classify a set of data; whereas, neural computing is a technique which uses a neural network (NN) model for classifying the input data [14]. With increasing demands of the problems, the NN approach for classifying pattern is becoming popular and new NN models have been developed in the process. The reason behind this fact is that, the statistical methods use certain set of assumptions and conditions to satisfy the solution. And these developed models cannot be applied directly to any such related problems [1]. This limitation of statistical methods has been reduced by the use of NN techniques. The first and foremost advantage of using NN techniques is that, NNs are nonlinear which let them adjust to the real world input data without any explicit specification or external conditions. Secondly, they are universal functional approximation models which can approximate any function with arbitrary accuracy [1,15,16]. NN techniques perform statistical analysis of the processed data to support the second advantage [1].

Significant progress has been made in recent years in the area of pattern recognition and neuro-computing. Feed-forward Multilayer Perceptrons (MLPs) are the most widely studied NN classifiers. A technique hybrid of fuzzy rules and NN, called neurofuzzy has been applied to model some real world problems such as medical diagnosis [17]. However, this technique has not been applied to pattern classification problem. A novel model was developed by Kang and Brown [18] for classifying data. This is an unsupervised NN, called as adaptive function neural

The final version of this paper has been published in "International Conference on Communication and Computing (ICC-2014)" http://www.elsevierst.com/conference_book_download_chapter.php?cbid=86#chapter41network and has no hidden layer in its architecture. A few examples of data classification applied in various fields have been listed below in the Table-1 with citation to corresponding works.

**Table 1** Classification in real world problems

| Application area | References |
|---|---|
| Bankruptcy prediction | [19,20,21,22,23] |
| Medical diagnosis | [4,5,17,24,25] |
| Handwritten signature recognition | [8,9,26,27] |
| Fingerprint recognition | [10,28,29] |
| System fault identification | [30] |
| Data classification | [31,32,33] |

Although the concept of Fuzzy MLP has been proposed by Mitra et al. [34] for classification and rule generation purposes, the model has not been applied successfully to pattern recognition or classification problems. This paper focuses on development of a novel Fuzzy MLP model for pattern recognition and classification. Performance of this model has been compared with that of the classical MLP model developed for the same problems. It should be noted that both fuzzy-MLP and classical MLP nets uses the concept of back propagation for learning patterns. The issues discussed in this paper can also apply to other neural network models.

The overall organization of this paper is as follows. A summary of preliminaries has been given in section 2 for refreshing the concept of MLP. Section 3 elaborates the proposed fuzzy-MLP model. Section 4 gives the experimental details and results followed by the conclusion section in section 5. A list of papers referred for this work has been given in reference section.

## 2. Preliminaries

### 2.1. The MLP architecture

The MLP is *n-layer* architecture, where *n≥3*. The layers are (i) input layer, (ii) output layers and (iii) hidden layer(s). The number of hidden layers may be increased or decreased based on the problem for which the model is developed. It is a feed-forward neural network, where non-linear elements, called neurons are arranged in layers. The data flow from the input layer to the output layer via hidden layer(s) [35]. The MLP net learns through supervised learning algorithms. A generalized delta rule (Δ-rule) is used for this process which has been given in pseudo-code section in the following section. The architecture of MLP is shown in Fig. 1. It should be noted that if there are *n* number of neurons in input layer, there can be a maximum of *2n+1* number of neurons in hidden layers, i.e. $m \leq 2n+1$.

The symbols used in the algorithms are given in the Table 2. It should be noted that, in this work, we have employed binary sigmoid as the activation function which is defined as $f(x) = 1/(1+exp(-x))$. Following equations are used for various updations mentioned in the above given piece of pseudo-code.

$$Z_{in} = b_1 + \sum_{i=1}^{n} X_i Vin_i \tag{1}$$

Where, $Vin_i$ represents the incoming input-hidden layer weight to the considered node for which net input $Z_{in}$ is being calculated. It should be noted that all the input, processing and output are carried out with decimal values only.

$$y_{in} = b_2 + \sum_{i=1}^{m} Z_i W_i \tag{2}$$

$$\Delta W = \{\alpha \times E \times f'(y_{in}) \times Z_i\} + (\mu \times \Delta W_{old}) \tag{3}$$

$$\Delta b_2 = \alpha \times E \times f'(y_{in}) \tag{4}$$

$$\Delta V = \{\alpha \times E \times f'(y_{in}) \times W \times f'(Z_{in}) \times X\} + (\mu \times \Delta V_{old}) \tag{5}$$

*Cite as:* **Dash, T., & Behera, H.S. (2014). A Fuzzy MLP Approach for Non-linear Pattern Classification.** *In Proc: K.R. Venugopal, S.C. Lingareddy (eds.) International Conference on Communication and Computing (ICC- 2014), Bangalore, India (June 12-14, 2014), Computer Networks and Security*, 314-323.

$$\Delta b_1 = \alpha \times E \times f'(y_{in}) \times W \times f'(Z_{in}) \tag{6}$$

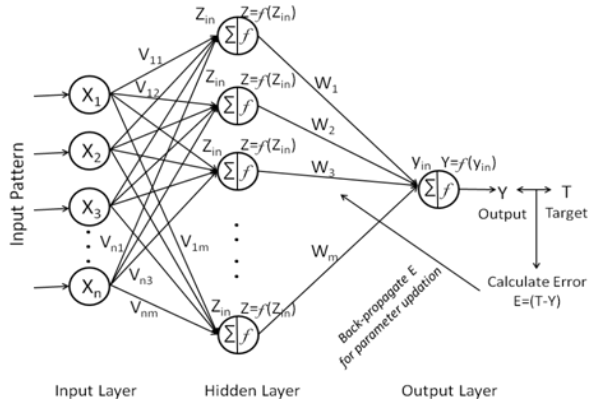

**Fig. 1** MLP architecture

**Table 2** Symbols used in algorithms

| Symbol | Name and Purpose |
|---|---|
| $W$ | Hidden-output layer weight (refer Fig. 1) |
| $\Delta W$ | change in weight, $W$ |
| $V$ | Input-hidden layer weight (refer Fig. 1) |
| $\Delta V$ | change in weight, $V$ |
| $b_1$ | bias value used in hidden layer |
| $\Delta b_1$ | change in $b_1$ |
| $b_2$ | bias value used in output layer |
| $\Delta b_2$ | change in $b_2$ |
| $\alpha$ | learning rate |
| $\mu$ | momentum factor |
| $f$ | activation function |

**Pseudo-code for training MLP:**

*Initialize* $W, V, b_1, b_2, \alpha, \mu$
*Set mean-squared-error, MSE = 0;*
*while* (termination criteria is not satisfied)
*do*
    *for* each training pattern
        *for* each hidden neuron
            *Calculate input to the hidden neuron ($Z_{in}$) using Equation-1*
            *Calculate activation of hidden neuron, $Z = f(Z_{in})$ (refer Fig. 1)*
        *end for*
        *Calculate net input to the output neuron ($y_{in}$) using Equation-2*
        *Calculate activation of output neuron, $Y = f(y_{in})$ (refer Fig. 1)*
        *Compute the error, $E = T – Y$; where T is the corresponding target*
        /*Back propagate the error to update weights*/
        *Calculate $\Delta W$ using Equation-3*
        *Update $b_2$ using Equation-4*
        *for* each hidden neuron
            *Calculate $\Delta V$ using Equation-5*
            *Update $b_1$ using Equation-6*
        *end for*
    *end for*
*end while*

## 3. Fuzzy MLP for pattern classification

In this work, an attempt has been made to introduce the concept of fuzzy membership with the MLP learning algorithm. The corresponding learning algorithm has been elaborated in the next section. We call the combined network formed thereof as Fuzzy-MLP. The membership function used in this model is the S-shaped (S-shaped) membership function or the spline-based membership function (MF) [36,37]. It should be noted that this type of MF is considered in this work because of its simplicity and it normalizes the input value to a certain range. However, other kind of MF can also be implemented for future researches in the area of pattern recognition and classification. This section is divided into three subsections as follows. Section-3.1 gives a brief description of S-shaped MF used in this work. Section-3.2 and 3.3 describes the proposed fuzzy-MLP architecture and corresponding learning algorithm respectively. Theoretical analysis of the proposed algorithm has been done in section-3.4.

### 3.1. Spline-based membership function (S-shaped MF)

A Spline-based MF is defined in equation-7 below. Where $a$, $b$ locate the extremes of sloped portion of the curve as shown in Fig. 2.

$$f(x,a,b) = \begin{cases} 0, & x < a \\ 2\left(\dfrac{x-a}{b-a}\right)^2, & a \leq x \leq \dfrac{a+b}{2} \\ 1 - 2\left(\dfrac{x-b}{b-a}\right)^2, & \dfrac{a+b}{2} \leq x \leq b \\ 1, & x > b \end{cases} \quad (7)$$

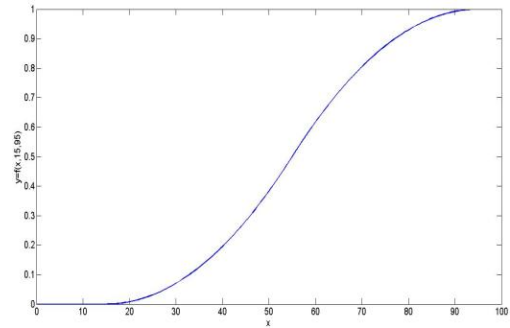

**Fig. 2.** S-shaped MF

### 3.2. Fuzzy-MLP network architecture

The proposed architecture is similar to that of the classical MLP architecture and has been shown in Fig.-3 below.

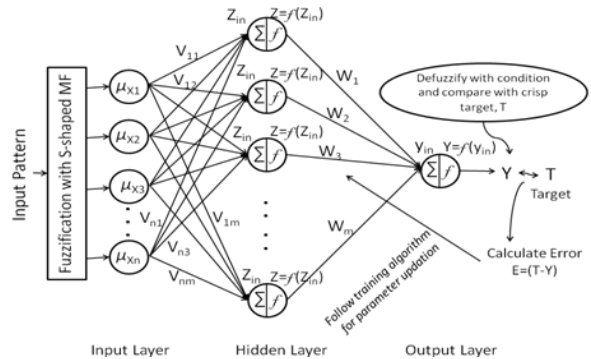

**Fig. 3.** Fuzzy-MLP architecture

### 3.3. Fuzzy-MLP learning pseudo-code

*initialize* W, V, $b_1$, $b_2$, α, μ, MSE=0
*fuzzify* the training input pattern using S-shaped MF
*input* the fuzzified values to the input layer neurons
*while* (termination criteria is not satisfied)
*do*
    *for* each training pattern
        *for* each hidden neuron
            Calculate input to the hidden neuron ($Z_{in}$)
            Calculate activation of hidden neuron, $Z = f(Z_{in})$ (refer Fig. 3)
        *end for*
    Calculate net input to the output neuron ($y_{in}$)

>
>         *Calculate activation of output neuron, Y = f($y_{in}$) (refer Fig. 3)*
>         *Y=Defuzzify the fuzzy output Y ($\mu_Y$)*
>         *Compute the error, E = T–Y; where T is the corresponding target*
>         */\*Back propagate the error to update weights*
>         *Calculate ΔW*
>         *Update $b_2$*
>         **for** *each hidden neuron*
>             *Calculate ΔV*
>             *Update $b_1$*
>         *end for*
>     *end for*
>     **update** *MSE = (MSE + $E^2$)/n*
> **end while**

### 3.4. Theoretical analysis of the proposed algorithm

In this section, time and space complexity analysis of the proposed algorithm has been carried out and represented briefly. In the proposed fuzzy-MLP model, number of input neurons is 'n'; number of hidden units is 'm' and number of output neuron is 1. Due to three layered architecture, the computation will be carried out in 4 major phases: forward (2-phases) and back-propagation (2-phases). The first forward computation is from input to hidden layer computation and it will occur in $O(n \times m)$ times. The next forward computation is from hidden to output layer computation and will take $O(m)$ times. Similarly, in the back propagation process, the steps output to hidden layer and hidden to input layer computations will occur in $O(m)$ and $O(m \times n)$ times respectively. So, the time complexity of the algorithm can be calculated as follows.

$$T(m,n) = 2 \times O(n \times m) + 2 \times O(m)$$
$$Or, \quad T(m,n) = O(n \times m); \text{ as } n \gg 1$$

In the proposed algorithm, one extra data matrix will be created which will store the fuzzified patterns for processing by the NN in the hidden layer. So, the overall space complexity is $O(n \times m)$. However, in the current era of computers, space complexity is not a problem anymore, as memory is abundantly available for computation. So, the result section is restricted to the time of processing only.

## 4. Experimental Details

The proposed algorithm has been evaluated with six public domain datasets from the University of California at Irvine (UCI) machine learning repository (*http://www.ics.uci.edu/mlearn/MLRepository.html*). The results have been compared with that obtained by using MLP. This section is divided into following subsections (i) description of dataset, (ii) experimental setup, and (iii) classification result.

### 4.1. Description of the datasets

    *IRIS:* This is most popular dataset for pattern classification. It is based on multivariate characteristics of flower plant species. Those are length and thickness of petals and sepals. The dataset contains three classes such as Iris Setosa, Iris Versicolor and Iris Virginica of 50 instances each. The dataset contains 150 instances and 5 attributes (4 predicting and 1 target). All the attribute values are real.

    *ABALONE:* The dataset contains physical characteristics of abalone such as sex, shell length, shell diameter, height, whole weight, shucked weight, viscera weight, shell weight and number of rings in the shell. The first seven characteristics determine number of rings in the shell and are treated as target class attribute in this work.

    *BREAST-CANCER-WISCONSIN (BCW):* This dataset is obtained from University of Wisconsin Hospitals, USA and contains attributes related to two major types of breast cancers; benign and malignant. The predicting attributes

are clump thickness, uniformity of cell size, shape, marginal adhesion, single epithelial cell size, bare nuclei, bland chromatin, normal nucleoli, and mitoses.

*GLASS:* The predicting attributes in this dataset are refractive index (RI), measurements of Na, Mg, Al, Si, K, Ca, Ba, Fe, which determine the quality of glass and for what purpose the glass is fit to be used. The types can be building-windows-float-processed, building-windows-non-float-processed, vehicle-windows-float-processed, vehicle-windows-non-float-processed, containers, tableware, and headlamps.

*SOYBEAN:* This is a small data set containing only 47 patterns but is having maximum number of attributes that is 36. This makes the NN to expand in the input as well as the hidden layers.

*WINE:* This dataset was generated as the results of a chemical analysis of wines grown in the same region in Italy but derived from three different cultivars. The analysis determined the quantities of 13 constituents found in each of the three types of wines. A few attributes are alcohol, malic acid, ash, magnesium etc.

**Table 3.** Dataset properties

| Dataset | Number of patterns | Number of attributes (including target attribute) | Number of target classes |
|---|---|---|---|
| Iris | 150 | 5 | 3 |
| Abalone | 4177 | 9 | 29 |
| Breast-cancer-Wisconsin (BCW) | 699 | 10 | 2 |
| Glass | 214 | 10 | 7 |
| Soybean | 47 | 36 | 4 |
| Wine | 178 | 14 | 3 |

*4.2. Experimental setup*

Both MLP and fuzzy-MLP algorithms are implemented using MATLAB R2010a environment installed in a personal computer having Windows 32-bit XP professional operating system and 2GB of main memory. The processor is Intel dual processor system (Intel Core2Duo) and each processor has equal computation speed of 2 GHz.

*4.3. Experimental results*

As mentioned earlier, the results obtained by employing both fuzzy-MLP and MLP algorithms are compared for each datasets. In this paper, convergence of MSE is considered for performance evaluation of pattern classification using both the algorithms. However, the learning parameters are varied from their minimum to maximum value and the results are noted. Number of nodes in the hidden node has set to 3/2 of the number of nodes in input layer; i.e. $3 \times n/2$. In this work, the learning rate ($\alpha$) is the most crucial parameter which affects the learning process of the NN. So, it is varied from 0 to 1 in the learning process and a result for each $\alpha$ is noted for each dataset as shown in next subsections. The typical values of $\alpha$ those were chosen are 0.05, 0.10, 0.25, 0.40, 0.55, 0.70, 0.85 and 0.99. The momentum factor ($\mu$) is set to 0.50 for all the experiments. As the fuzzy-MLP is converging very much faster than the MLP net, the number of epoch is set to 100. A table has been maintained for each datasets to show the minimum MSE obtained by MLP and fuzzy-MLP. The table also shows the time consumed by both the algorithms for the process of pattern classification. To check the efficiency of fuzzy-MLP algorithm over classical MLP model, a factor called convergence gain ($C_g$) has been introduced and is defined in equation-8.

$$C_g = \frac{Min\_MSE_{MLP} - Min\_MSE_{Fuzzy-MLP}}{Min\_MSE_{MLP}} \quad (8)$$

*4.3.1. Case-1 (IRIS dataset)*

The training performance of the fuzzy-MLP and MLP algorithms are evaluated with Iris dataset. The result summary has been given in Table 5. It should be noted that the simulation time presented in the table is an average of 5 executions. As number of input attributes for Iris dataset is 4, the number of neurons in hidden layer is set to 6. Results are shown for 100 epochs and $\mu$ is set to 0.50. It is clear from the above table that when $\alpha$ is set to 0.99, the propose fuzzy-MLP algorithm is converging to a minimum error of 0.016 (approx.) within 100 epochs only where as the MLP algorithm is converging only to 1.67 (approx.) with this number of epochs. The convergence gain is approximately 99% when $\alpha$ is set to 0.99. However, the overall performance of the proposed algorithm is better than the classical MLP algorithm when the convergence gain is considered. The simulation time of the proposed algorithm is also lesser than that of MLP algorithm in almost all $\alpha$ settings. A plot has been given in Fig. 4 by taking number of epoch in X-axis and MSE in Y-axis to illustrate the convergence of error in fuzzy-MLP algorithm for each $\alpha$. The Fig. also shows that $\alpha = 0.99$ is the best value for pattern classification in Iris dataset. However, $\alpha$ can also be set to 0.85 for which the result of convergence is approximately equal to that in $\alpha = 0.99$.

*4.3.2. Case-2 (ABALONE dataset)*

The result summary for Abalone dataset has been given in Table 6. Here number of hidden units is equal to 12.Unlike results in Iris dataset, the minimum MSE obtained by MLP and fuzzy-MLP algorithms are having small difference of 0.39. The gain also shows that the later one is not showing better efficiency as compared to that in the Iris dataset. The plot given in Fig. 5 shows the convergence of error when $\alpha$ is set to the different values. It can be seen from table-5 and the above given plot that the convergence of error is hardly depending on the learning rate for the Abalone dataset. However, with $\alpha$=0.05 and 0.10, the error is converging to its minimum after 15 epochs, where as for other $\alpha$ values it's getting to its minimum within 10 epochs.

*4.3.3. Case-3 (BCW dataset)*

The result summary for BCW dataset has been given in Table 7. Here number of hidden units is equal to 14. The fuzzy-MLP algorithm outperforms the classical MLP algorithm with a convergence gain factor of approximately 98%. As the BCW dataset is large, it can be proposed that the proposed algorithm can be applied to larger datasets in real life scenarios. A plot has been given in Fig. 6 shows the convergence of error when $\alpha$ is set to the different values.

*4.3.4. Case-4 (GLASS dataset)*

The result summary for Glass dataset has been given in Table 8. Here number of hidden units is set to 15, as the number of input attributes is 10. It can be seen from the table that the maximum gain obtained is 99% when $\alpha$ is set within a range of 0.25-0.99. The plot given in Fig. 7 shows the convergence of error when $\alpha$ is set to the different values between 0 and 1.

*4.3.5. Case-5 (SOYBEAN dataset)*

The result summary for Soybean dataset has been given in Table 9. Here number of hidden units is equal to 51. It can be noted that the Soybean dataset is the dataset is having maximum input characteristic, i.e. 34. And the average gain obtained is approximately 95%. Therefore, the proposed algorithm can be applied to problems with larger predicting attributes. The plot given in Fig. 8 shows the convergence of error when $\alpha$ is set to the different values. It can be seen from Fig. 8 that, the MSE obtained by the employing proposed fuzzy-MLP algorithm are converging to their corresponding minimum with different $\alpha$ after 25 epochs. The plots with *$\alpha$=0.05 and $\alpha$=0.25* are not converging within 100 epochs. However, observing the plots, the converging is inversely proportional to $\alpha$. That means the result with maximum $\alpha$ is converging with minimum epoch and result with minimum $\alpha$ is converging with more number of epochs.

*4.3.6. Case-6 (WINE dataset)*

The result summary for Wine dataset has been given in Table 10. Here number of hidden units is equal to 18. Observing the MSEs obtained by fuzzy-MLP algorithm with α set to 0.05 and 0.10 there is a difference of 0.27. The plot given in Fig. 9 shows the convergence of error when $\alpha$ is set to the different values. The convergence obtained with the Wine dataset is similar to that of Glass dataset where the MSE plot is gradually increasing after 40 epochs.

**Table 5** Performance of fuzzy-MLP for Iris dataset

| α | Minimum MSE | | Convergence Gain ($C_g$) | Simulation Time (sec) | | | MLP | | | | MLP |
|---|---|---|---|---|---|---|---|---|---|---|---|
| | MLP | Fuzzy- | | MLP | Fuzzy- | 0.05 | 1.66686 | 0.07183 | 0.9569 | 62.82 | 55.82 |
| | | | | | | 0.10 | 1.66676 | 0.04323 | 0.9741 | 62.77 | 55.79 |
| | | | | | | 0.25 | 1.66670 | 0.03665 | 0.9780 | 56.06 | 50.55 |

*Cite as:* **Dash, T., & Behera, H.S. (2014). A Fuzzy MLP Approach for Non-linear Pattern Classification.** *In Proc: K.R. Venugopal, S.C. Lingareddy (eds.) International Conference on Communication and Computing (ICC- 2014), Bangalore, India (June 12-14, 2014), Computer Networks and Security*, 314-323.

| 0.40 | 1.66669 | 0.02718 | 0.9837 | 57.05 | 56.70 |
| 0.55 | 1.66668 | 0.02175 | 0.9870 | 58.96 | 47.94 |
| 0.70 | 1.66668 | 0.01873 | 0.9888 | 55.18 | 56.11 |
| 0.85 | 1.66667 | 0.01687 | 0.9899 | 60.62 | 49.46 |
| 0.99 | 1.66667 | 0.01566 | 0.9906 | 60.81 | 46.78 |

**Table 6** Performance of fuzzy-MLP for Abalone dataset

| α | Minimum MSE | | Convergence Gain ($C_g$) | Simulation Time (sec) | |
|---|---|---|---|---|---|
| | MLP | Fuzzy-MLP | | MLP | Fuzzy-MLP |
| 0.05 | 0.5105 | 0.1045 | 0.7952 | 1832.82 | 1844.7 |
| 0.10 | 0.5024 | 0.1034 | 0.7942 | 838.68 | 885.39 |
| 0.25 | 0.4952 | 0.1035 | 0.7909 | 866.88 | 863.60 |
| 0.40 | 0.4894 | 0.1035 | 0.7885 | 846.67 | 841.34 |
| 0.55 | 0.4843 | 0.1035 | 0.7862 | 842.05 | 838.90 |
| 0.70 | 0.4815 | 0.1036 | 0.7848 | 841.62 | 840.23 |
| 0.85 | 0.4794 | 0.1037 | 0.7836 | 834.07 | 847.16 |
| 0.99 | 0.4780 | 0.10394 | 0.7826 | 824.41 | 843.72 |

**Table 7** Performance of fuzzy-MLP for BCW dataset

| α | Minimum MSE | | Convergence Gain ($C_g$) | Simulation Time (sec) | |
|---|---|---|---|---|---|
| | MLP | Fuzzy-MLP | | MLP | Fuzzy-MLP |
| 0.05 | 3.75824 | 0.04232 | 0.9887 | 152.17 | 156.42 |
| 0.10 | 3.75823 | 0.04122 | 0.9890 | 154.13 | 158.57 |
| 0.25 | 3.75823 | 0.04128 | 0.9890 | 156.24 | 149.07 |
| 0.40 | 3.75822 | 0.04150 | 0.9890 | 148.39 | 151.29 |
| 0.55 | 3.75822 | 0.04169 | 0.9889 | 156.81 | 150.67 |
| 0.70 | 3.75822 | 0.04183 | 0.9889 | 147.51 | 147.36 |
| 0.85 | 3.75822 | 0.04195 | 0.9888 | 149.75 | 148.83 |
| 0.99 | 3.75822 | 0.04207 | 0.9888 | 153.26 | 153.62 |

**Table 8** Performance of fuzzy-MLP for Glass dataset

| α | Minimum MSE | | Convergence Gain ($C_g$) | Simulation Time (sec) | |
|---|---|---|---|---|---|
| | MLP | Fuzzy-MLP | | MLP | Fuzzy-MLP |
| 0.05 | 7.57481 | 0.12105 | 0.9840 | 44.92 | 53.37 |
| 0.10 | 7.57479 | 0.09537 | 0.9874 | 44.53 | 48.71 |
| 0.25 | 7.57478 | 0.07441 | 0.9902 | 43.56 | 41.70 |
| 0.40 | 7.57476 | 0.06987 | 0.9908 | 43.81 | 41.64 |
| 0.55 | 7.57474 | 0.06626 | 0.9913 | 42.39 | 42.81 |
| 0.70 | 7.57472 | 0.06396 | 0.9916 | 42.30 | 45.66 |
| 0.85 | 7.57471 | 0.06217 | 0.9918 | 44.64 | 43.78 |
| 0.99 | 7.57471 | 0.06067 | 0.9920 | 42.69 | 41.59 |

**Table 9** Performance of fuzzy-MLP for Soybean dataset

| α | Minimum MSE | | Convergence Gain ($C_g$) | Simulation Time (sec) | |
|---|---|---|---|---|---|
| | MLP | Fuzzy-MLP | | MLP | Fuzzy-MLP |
| 0.05 | 4.31924 | 0.35186 | 0.9185 | 10.68 | 11.65 |
| 0.10 | 4.31922 | 0.35183 | 0.9185 | 12.06 | 12.08 |
| 0.25 | 4.31902 | 0.35131 | 0.9187 | 11.09 | 11.55 |
| 0.40 | 4.31918 | 0.10119 | 0.9766 | 10.99 | 9.43 |
| 0.55 | 4.31918 | 0.03035 | 0.9930 | 9.61 | 9.20 |
| 0.70 | 4.31917 | 0.20958 | 0.9515 | 9.38 | 9.83 |
| 0.85 | 4.31917 | 0.02967 | 0.9931 | 10.17 | 10.19 |
| 0.99 | 4.31916 | 0.02975 | 0.9931 | 10.38 | 10.16 |

**Table 10** Performance of fuzzy-MLP for Wine dataset

| α | Minimum MSE | | Convergence Gain ($C_g$) | Simulation Time (sec) | |
|---|---|---|---|---|---|
| | MLP | Fuzzy-MLP | | MLP | Fuzzy-MLP |
| 0.05 | 1.47754 | 0.36559 | 0.7526 | 37.71 | 36.65 |
| 0.10 | 1.47754 | 0.08701 | 0.9411 | 37.73 | 42.44 |
| 0.25 | 1.47753 | 0.06913 | 0.9532 | 37.40 | 35.42 |
| 0.40 | 1.47753 | 0.06440 | 0.9564 | 35.91 | 40.93 |
| 0.55 | 1.47753 | 0.06271 | 0.9576 | 35.34 | 36.01 |
| 0.70 | 1.47753 | 0.06352 | 0.9570 | 36.99 | 36.90 |
| 0.85 | 1.47753 | 0.06679 | 0.9548 | 39.09 | 39.30 |
| 0.99 | 1.47753 | 0.06226 | 0.9579 | 39.81 | 40.36 |

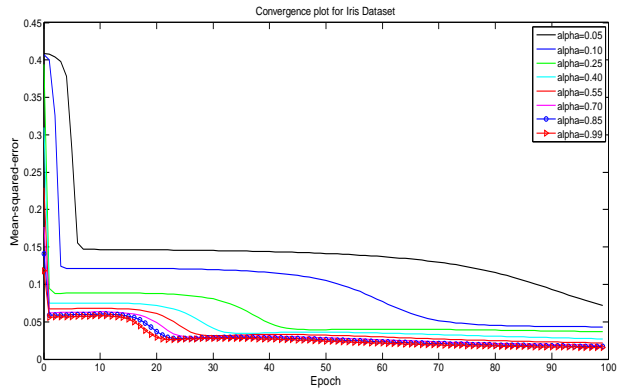

**Fig.4.** Convergence plot obtained by employing Fuzzy MLP algorithm for Iris dataset

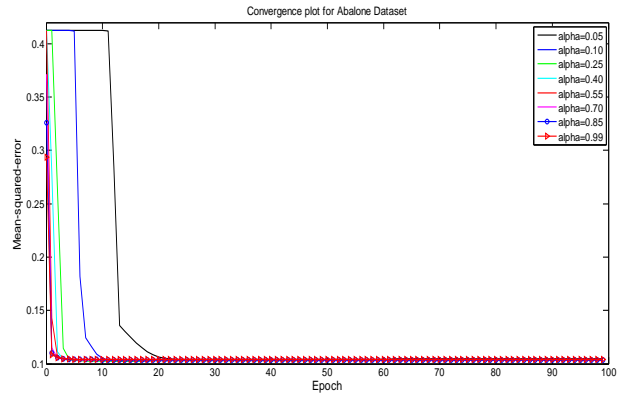

**Fig. 5.** Convergence plot obtained by employing fuzzy-MLP algorithm for Abalone dataset

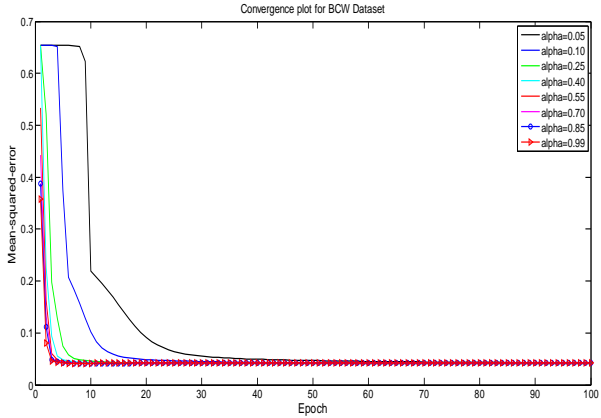

**Fig.6.** Convergence plot obtained by employing fuzzy-MLP algorithm for BCW dataset

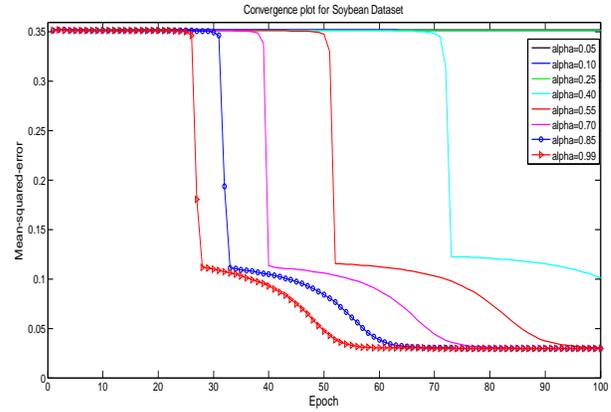

**Fig. 8.** Convergence plot obtained by employing fuzzy-MLP algorithm for Soybean dataset

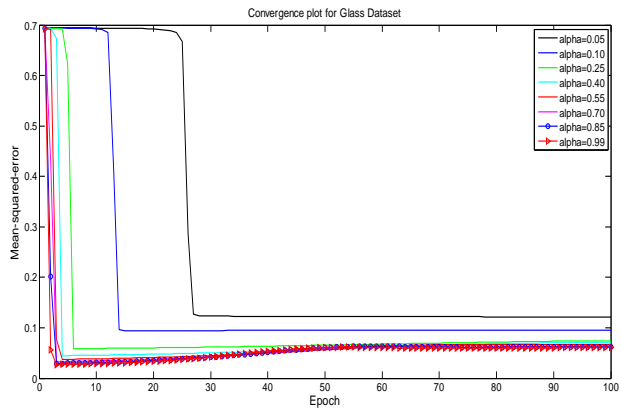

**Fig. 7.** Convergence plot obtained by employing fuzzy-MLP algorithm for Glass dataset

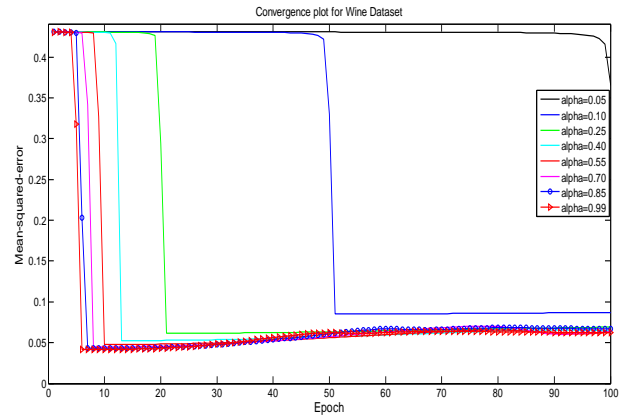

**Fig. 9.** Convergence plot obtained by employing fuzzy-MLP algorithm for Wine dataset

For all the datasets, the proposed algorithm results a higher gain when compared with the MLP model. However, as the MSE varies with $\alpha$ value, it will be interesting to analyze a plot for the gain against each $\alpha$. A plot has been given in Fig. 10 by $\alpha$-value in X-axis and average gain (average $C_g$) in Y-axis. It should be noted that average gain against an $\alpha$ is obtained by calculating the average of all the gains obtained with the six datasets used in this work. Fig. 10 clearly reveals that best gain, which is close to 95%, is obtained when the parameter $\alpha$ is tuned to 0.55, 0.85 and 0.99. So, it should be noted that for any datasets including the six UCI datasets used in this work, the proposed NN model will achieve a higher convergence gain. However, it will also be important to check if the simulation time is affected by $\alpha$. Therefore, a plot has been given in Fig. 11 to show a comparative characteristic of average simulation time for different $\alpha$ values. As mentioned in previous section, the simulation time is obtained by calculating mean of simulation time obtained with five consecutive executions to make the result error free. In this plot, $\alpha$ has been taken in X-axis and average simulation time is shown in Y-axis.

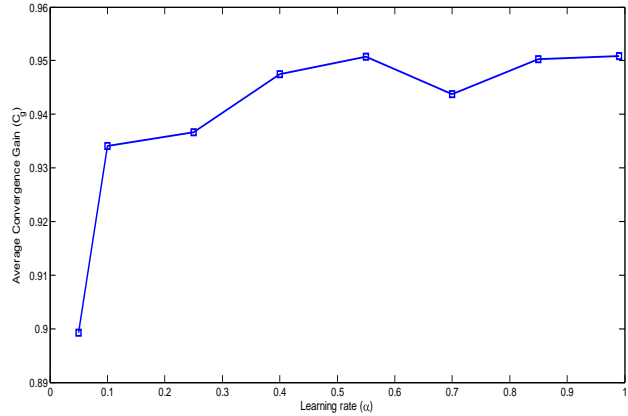

**Fig. 10.** Average gain obtained against different learning rate ($\alpha$)

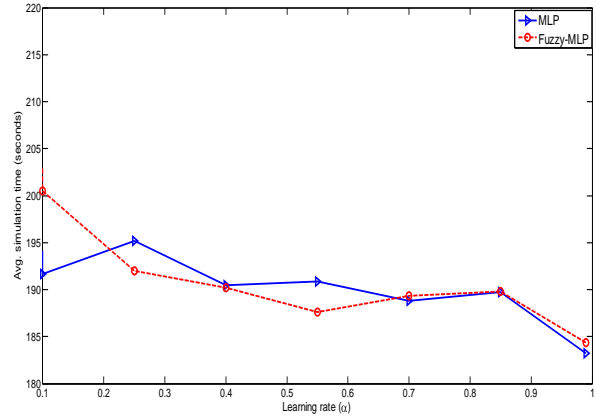

**Fig. 11.** Average simulation time obtained against different learning rate ($\alpha$)

In the previous discussion, it is seen that when $\alpha$ is tuned to 0.55, 0.85 or 0.99, the proposed model obtains a maximum of 95% convergence gain. So, it will be wise to check the simulation time against these three $\alpha$ values only. Fig. 11 shows, for $\alpha=0.55$, the proposed model is trained within 190 seconds whereas the MLP model is taking more than 190 seconds for the purpose. For rest two $\alpha$ values, the simulation is approximately equal to that obtained by using the MLP model and it is more than that obtained by setting $\alpha$ to 0.55. Therefore, it could be concluded that the best $\alpha$ is 0.55 for all the future tests.

### 4.3.7. Statistical analysis of result (ANOVA)

ANOVA test on the results has been conducted carefully to see if there is any difference between the groups on the resulted data. The t-test result using ANOVA has been given in Table-11 below. It should be noted that the test has been carried out using the SPSS software package [38].

**Table 11** t-test on the result data

| Variables | t-value | Mean Difference | 95% Confidence Interval of the Difference | |
|---|---|---|---|---|
| | | | Lower | Upper |
| $MSE_{MLP}$ | 9.285 | 3.24734596 | 2.5433497 | 3.9513422 |
| $MSE_{Fuzzy-MLP}$ | 6.970 | .09052234 | .0643786 | .1166661 |
| $C_g$ | 81.684 | .9387638 | .915630 | .961897 |
| $T_{MSE}$ | 3.947 | 214.68277 | 105.2108 | 324.1548 |
| $T_{Fuzzy-MSE}$ | 3.925 | 215.65191 | 105.0514 | 326.2524 |

## 5. Conclusion

The fuzzy-MLP for pattern classification has been developed. The input patterns are fuzzified by using spline (S-shaped) membership function and then input to the MLP model. The results obtained shows that, the proposed model converges to its minimum MSE within 100 epochs and achieves a convergence gain of 93%. The proposed algorithm outperforms MLP for all the six UCI datasets used in this work. As future work, it will be advantageous to optimize the network with an optimization algorithm.

*Cite as:* **Dash, T., & Behera, H.S. (2014). A Fuzzy MLP Approach for Non-linear Pattern Classification.** *In Proc: K.R. Venugopal, S.C. Lingareddy (eds.) International Conference on Communication and Computing (ICC- 2014), Bangalore, India (June 12-14, 2014), Computer Networks and Security*, 314-323.